\documentclass[sigconf]{acmart}

\usepackage[normalem]{ulem}
\usepackage{enumitem}
\usepackage{algorithm}
\usepackage{algpseudocode}
\usepackage{multirow}

\PassOptionsToPackage{prologue,dvipsnames}{xcolor}
\PassOptionsToPackage{prologue,dvipsnames}{xcolor}
\usepackage{threeparttable}

\AtBeginDocument{%
  \providecommand\BibTeX{{%
    \normalfont B\kern-0.5em{\scshape i\kern-0.25em b}\kern-0.8em\TeX}}}

\setcopyright{acmcopyright}
\copyrightyear{2023}
\acmYear{2023}
\acmDOI{10.1145/3589132.3625607}

\acmConference[]{SIGSPATIAL ’23}{November 13–16,
  2023}{Hamburg, Germany}
\acmBooktitle{The 31th International Conference on Advances in Geographic Information Systems (SIGSPATIAL ’23), November 13–16, 2023,Hamburg, Germany} 
\acmPrice{15.00}
\acmISBN{979-8-4007-0168-9/23/11}

\begin{document}


\title{Explainable Trajectory Representation  \protect\\ through Dictionary Learning }


    

\author{Yuanbo Tang}
\affiliation{%
  \institution{Tsinghua University}
  \streetaddress{}
  \city{}
  \state{}
  \country{}}
  \email{tyb22@mails.tsinghua.edu.cn}

  \author{Zhiyuan Peng}
\affiliation{%
  \institution{Tsinghua University}
  \streetaddress{}
  \city{}
  \state{}
  \country{}}
\email{ pengzy21@mails.tsinghua.edu.cn}

  \author{Yang Li*}
\affiliation{%
  \institution{Tsinghua University}
  \streetaddress{}
  \city{}
  \state{}
  \country{}}
  \email{yangli@sz.tsinghua.edu.cn}

\newcommand\outline[1]{}

\newcommand{\tobechange}[1]{\textcolor{red}{\uline{#1}}}
\newcommand{\changed}[1]{\textcolor{blue}{\uline{#1}}}

\newcommand\normalText[1]{\textcolor[RGB]{18,220,168}{#1}}
\renewcommand{\thefootnote}{}

\begin{abstract}
 
Trajectory representation learning on a network 
enhances our understanding of
vehicular traffic patterns and benefits numerous downstream applications. Existing approaches using classic machine learning or deep learning embed trajectories as dense vectors, which lack interpretability and are inefficient to store and analyze in downstream tasks. In this paper, an explainable trajectory representation learning framework through dictionary learning is proposed.  Given a collection of trajectories on a network, it extracts a compact dictionary of commonly used subpaths called “pathlets”, which optimally reconstruct each trajectory by simple concatenations. The resulting representation is naturally sparse and encodes strong spatial semantics. Theoretical analysis of our proposed algorithm is conducted to provide a probabilistic bound on the estimation error of the optimal dictionary.   A hierarchical dictionary learning scheme is also proposed to ensure the algorithm’s scalability on large networks, leading to a multi-scale trajectory representation.  Our framework is evaluated on two large-scale real-world taxi datasets. Compared to previous work, the dictionary learned by our method is more compact and has better reconstruction rate for new trajectories.  
We also demonstrate the promising performance of this method in downstream tasks including trip time prediction task and data compression.
\vspace{-0.25cm}
\end{abstract}




  \vspace{-0.2cm}

\maketitle
  \vspace{-0.2cm}
\section{Introduction}

The 
development of information technology and the 
widespread use of mobile devices have produced a large amount of GPS trajectory data. Raw trajectory data typically appears as variable-length ordered sequences, which cannot be directly input into common data mining algorithms. Trajectory representation learning, which means transforming a trajectory into an embedding vector, can standardize trajectory data, extract valuable information from redundant original data, and benefit various downstream tasks including trajectory compression, trip time estimation \cite{cao_han-lin_survey_2021}. 

Recently, various deep learning based models for trajectory representation learning has been developed.
For example, Yang et al. \cite{al_t3s_2021} introduced a model based on self-attention (T3S) that automatically adjusts the importance of spatial and structure information for different similarity measures. And they showed the effectiveness for trajectory similarity computation. 
In addition, in \cite{liang_nettraj_2022} the authors proposed a trajectory encoder-decoder network based on graph attention mechanism to obtain trajectory embedding and evaluate in vehicle trajectories prediction task.
Before the emergence of these deep learning based methods, researchers also attempted to explore this field using traditional algorithms, 
including \cite{zygouras_corridor_2018}, wherein the authors introduce a pipelined algorithm that extract frequent underlying paths called corridor from trajectories and evaluate it using Minimum Description Length (MDL) score. Besides that, Zou et al. \cite{zou_cluster_2014} extracted middle level features from trajectories for clustering using a cluster specific Latent Dirichlet Allocation Model. 
 However, the representations generated by previous methods are usually dense vector whose dimensions lack semantic meanings.  As a result, it is difficult to interpret the learned representation.
\vspace{-0.4cm}
\begin{figure}[h]
\centering
\includegraphics[width=0.45\textwidth]{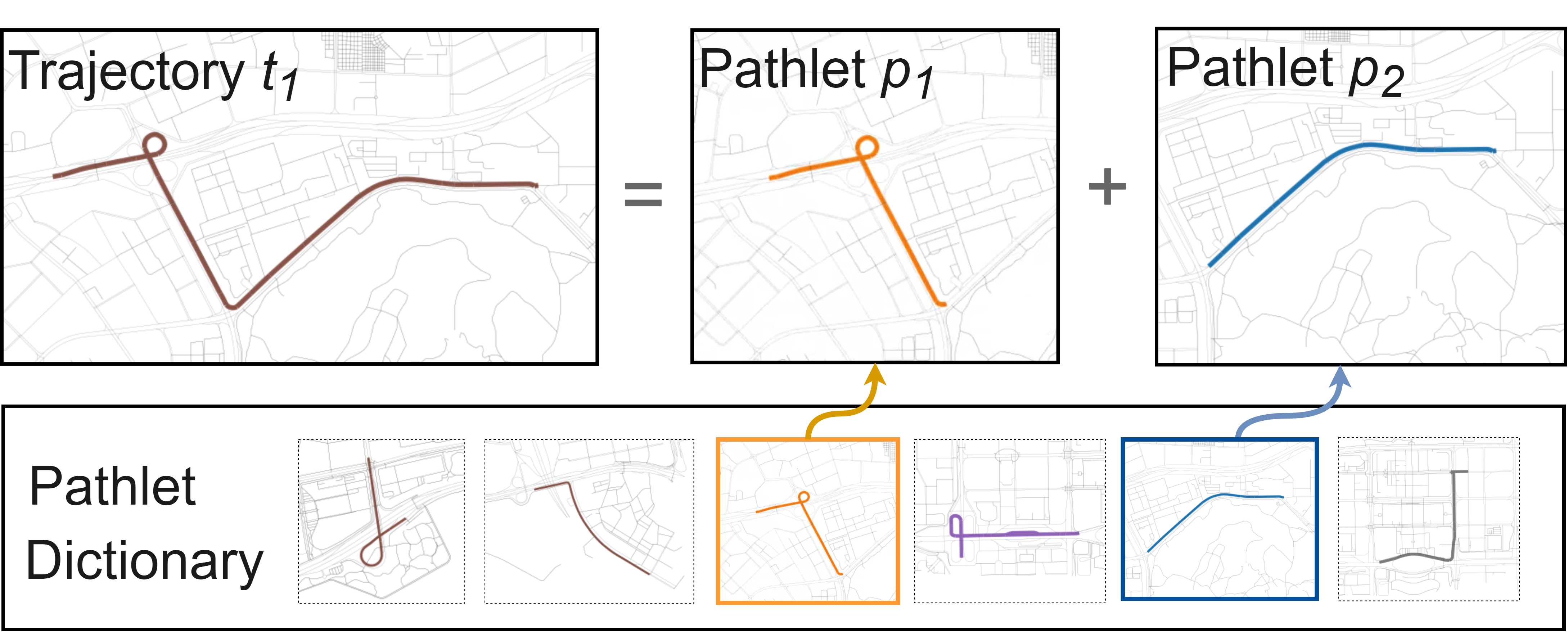}
\captionsetup{skip=5pt}
\caption{\label{fig:pic1} Illustration of pathlet learning: A pathlet dictionary is learned from dataset and each trajectory can be represented by concatenating pathlets chosen from this dictionary.}
\vspace{-0.2cm}
\end{figure}
 
 
 In this paper, we introduce an explainable trajectory representation method through dictionary learning for trajectories on a network. The network is usually  a road map for vehicle trajectories or a grid network for unstructured trajectories, on which trajectory can be projected using map matching  \cite{newson_hidden_2009}.
 \footnotetext{Full text version of this paper is available at  github.com/tangyuanbo1/pathlet-learning
 }
 The basic idea is demonstrated in Figure \ref{fig:pic1}.
Given a collection of trajectories on a network, it extracts a compact dictionary of commonly used subpaths called “pathlets”.
 Each trajectory can then be reconstructed by concatenating pathlets from the dictionary, similar to the process of constructing a sentence by assembling a group of words.  
 The resulting trajectory representation is a sparse binary vector, where each dimension corresponds to a pathlet in the learned dictionary and each binary variable indicates whether the corresponding pathlet is used to reconstruct the trajectory. 
 Such design is motivated by the observation that people's travel behavior exhibits remarkable regularity, enabling us to reconstruct majority of trajectories using a small set of movement patterns. 
 
 The pathlet representation of trajectories was first explored by Chen et al. \cite{chen_pathlet_2013}, who formulate the pathlet learning problem as a combinatorial optimization problem. Solved approximately using dynamic programming, the original formulation is costly to compute and lacks theoretical guarantee.  
 We propose an algorithm using a novel dictioanry learning formulation that provides better  optimality and scalability  for large trajectory datasets.
Specifically, in  our formulation, the objective function minimizes the size of the pathlet dictionary and the average number of pathlets required to reconstruct each trajectory at the same time. We propose an efficient solution to this integer programming problem, by first solving its relaxed version and find the integer solution using randomized rounding. To ensure the scalability to large-scale road networks, we further propose a hierarchical representation scheme that compute  pathlets of different granularity in multi-scale spatial partition of the map. This algorithm is evaluated on two real-world taxi datasets and some frequent mobility patterns are visualized. 
We also demonstrate the promising performance of this method in downstream tasks. For example, our method outperforms neural-network based methods by 4.7\% in prediction accuracy on trip time prediction.

  \vspace{-0.15cm}
\section{Preliminary }

\noindent\textbf{Terminology.}  
Given a dataset $T$ and a roadmap that can be formed as a directed graph $G=(E,V)$, a trajectory $t\in T$ 
is defined as a sequence of edges $e$ on $G$. 
For each $t$, a path $p$ on $G$ is a candidate pathlet if $p$ is a subpath of $t$. 
We denote the set of all candidate pathlets traversed by T  as $\overline{P}$.    



Given a pathlet dictionary $P$ and a trajectory $t$, $P_{sub}$ is a subset of $P$ so that $t$ can be represented 
 by concatenating $p \in P_{sub}$. This process is denoted by $t=c(P_{sub})$. Furthermore,
 the representation cost $rc(t, P)$ refers to the  minimal number of elements required to represent $t$ , which is defined as:
$rc(t, {P})=\min \limits_{{{P}_{ {sub }} \subset {P}, t=c\left({P}_{ {sub }}\right)}} \left|{P}_{ {sub }}\right|$


\noindent\textbf{Problem definition.}
The goal is to find an optimal dictionary $P$ that minimizes the following two objectives at the same time: 
1) the size of the dictionary,  as a smaller dictionary contains less redundant information and is therefore more desirable.
2) the average number of elements required to reconstruct trajectories. We use hyperparameter $\lambda$ to control the trade-off between these two objectives.  
Therefore, similar to \cite{chen_pathlet_2013}, in this paper the pathlet dictionary learning problem is defined as:
\begin{align} 
&\underset{P\subset \overline{P}}{\min}  \ \ size({P})+\lambda * \sum_{t\in T} rc(t, P) 
\\ &s.t.\forall t\in{T},\exists{P}_{\mathrm{sub}}\subset{P}:t=c({P}_{\mathrm{sub}})
\end{align}



  \vspace{-0.2cm}
\section{Methodology}
\label{sec:med}

 \subsection{Problem Formulation}

To formulate the problem  defined above using vector notations, we introduce three matrices $M$, $D$, $R$ to record the cover relationship among trajectories, edges, and candidate pathlets respectively. 
Matrix $M$ has dimensions of $|E|$ by $|T|$, where each element $M_{i,j}$ 
is equal to 1 when the i-th trajectory passes through the j-th edge and 0 otherwise. Matrix $D$ with a size of $|E|\times|\overline{P}|$ is constructed in the same way for the relationship between all candidate pathlets and edges. Similarly, matrix $R$ is a $|\overline{P}|\times|T|$ decision matrix, each entry $R_{i,j}=1$ if $p_i$ is used to represent $t_j$, and  $R_{i,j}=0$ otherwise. 
 \footnote{
In practice, $|\overline{P}|$ can be quite large. We pre-filter out less frequently used candidates to alleviate computational burden. Please refer to Appendix B for details.}
Based on these definition, the problem can be formulated as follows:
\vspace{-0.15cm}
\[ \min\limits_{{R_{i,j} \in \{0,1\}}} \ \ C(R)=\sum_{i=1}^{|\overline{P}|}{max(R_{i,:})}+\lambda*\sum_{i=1}^{|\overline{P}|}\sum_{j=1}^{|T|}|R_{i,j}| \]
\vspace{-0.35cm}
$$s.t. DR= M$$

Here $max(R_{i,:})$ refers to the maximum value of $i$ th row of $R$, which is equal to 1 if any trajectory utilizes $p_i$ to represent itself. In other words, $max(R_{i,:})=1$  means that candidate pathlet $p_i$ is selected as an element of the dictionary.  We reuse the notation $P$ to represent the matrix form of the dictionary, which is a submatrix formed by selected columns of $D$, 
$P = D[:,  \{i\mid max(R_i,:)=1\} ]$
and therefore $size({P})=\sum_{i=1}^{|\overline{P}|}{max(R_{i,:})}$.  The constraint $DR= M$ corresponds to the setting that each trajectory should be reconstructed using pathlets. In this optimization problem, the dictionary and the assignment relationship will be optimized at the same time.
It is worth noting that the pathlet learning problem described above is NP-hard in most cases. Therefore, an effective algorithm is required to obtain good approximated solutions.
  \vspace{-0.15cm}
 \subsection{Pathlet  Dictionary Learning with Randomized Rounding}
The proposed  algorithm consists of two main steps. 
Firstly, we relax the binary constraint, which transforms the original optimization problem into a convex optimization problem.  Therefore, the global optimal solution $R^*$ can be found easily by the projected gradient descent algorithm. Then a randomized rounding step is carried out to obtain the final solution $R^r$. The whole procedure is shown in the following pseudocode of Algorithm 1.
\vspace{-0.15cm}
\begin{algorithm}[h]
  \caption{Pathlet dictionary learning by randomized rounding} 
  \label{alg::conjugateGradient}
  \begin{algorithmic}[1]
    \Require
      $M$: trajectory matrix;
      $D$: pathlet matrix;
      $R_0$: initial solution;
      $\epsilon,\theta$: hyperparameters;
      $C$: objective function;

    \Ensure
      Optimal binary matrix $R^{r}$
     
     \State\noindent \# Step1, we compute the fractional solution $R^*$.
    \State initial $R_0=\mathbf{0}$;
    \Repeat
      \State compute gradient directions $g_k=\bigtriangledown C(R_k)$;
      \State update the decision matrix $R_{k+1}=R_k-\alpha g_k$;
      \State clip the result to make sure $0\le R_k \le 1$;
    \Until{($|C(R_k)-C(R_{k-1})|<\epsilon$)}\\
    \noindent \#Step2, we compute rounded solution $R^r$ based on $R^*$.
     \State Sample $R^r$ with $P(R^r_{i,j}=1)=min(1,\theta R^*_{i,j})$
  \end{algorithmic}
\end{algorithm}

\noindent\textbf{Probabilistic Bound.}
Given constant matrices ($M,D$)  and hyperparameters ($\theta,\lambda$) ,  
 We claim that the final solution $R^r$ satisfies
\[
 P[C(R^r)\le 2\theta \frac{\lambda+1}{\lambda}C(R^*)\ and\ DR^r\ge M] \ge \frac{1}{2}-|T|e^{-\theta}
\]
\noindent{This inequality means that the probablity that a solution with low cost can be found and all trajectories will be covered at the same time is lower bounded by a positive constant. 
Therefore, we can repeat the randomized rounding process to  get a series of $\{R^r_1,R^r_2...\}$ until find a satisfactory solution. } The proof can be found in appendix part A.

  \vspace{-0.15cm}
\subsection{Hierarchical Pathlet Learning}
\label{sec:Hierarchical}
Candidate pathlet space consists of all  segments of trajectories from dataset, whose size is usually huge in real-world dataset and make it time-consuming to get the solution. On the other hand, 
 Multi-scale dictionaries of pathlets and trajectory representations  can help people gain a deeper understanding of traffic characteristic.
To enhance the scalability of the original algorithm, we introduce a hierarchical method called "pathlet of pathlets" to reduce the computation complexity and generate multi-scale trajectory representations. 



Specificly, we first partition the roadmap into  different levels of granularity using axis-aligned binary space partitioning. Starting from the bottom of the partition tree, we compute the $k$-th level pathlet dictionary  $P_k$  as the union of dictionaries computed in all $k$-th level cells. Next, we use the $k$-th level pathlet representation of each trajectory as the input, and compute the ($k-1$)-th level  pathlet  dictionaries. 
This iterative process can be repeated to generate multi-scale pathlets that capture movement patterns.

\vspace{-0.35cm}
\begin{figure}[h]
\centering
\includegraphics[width=0.48\textwidth]{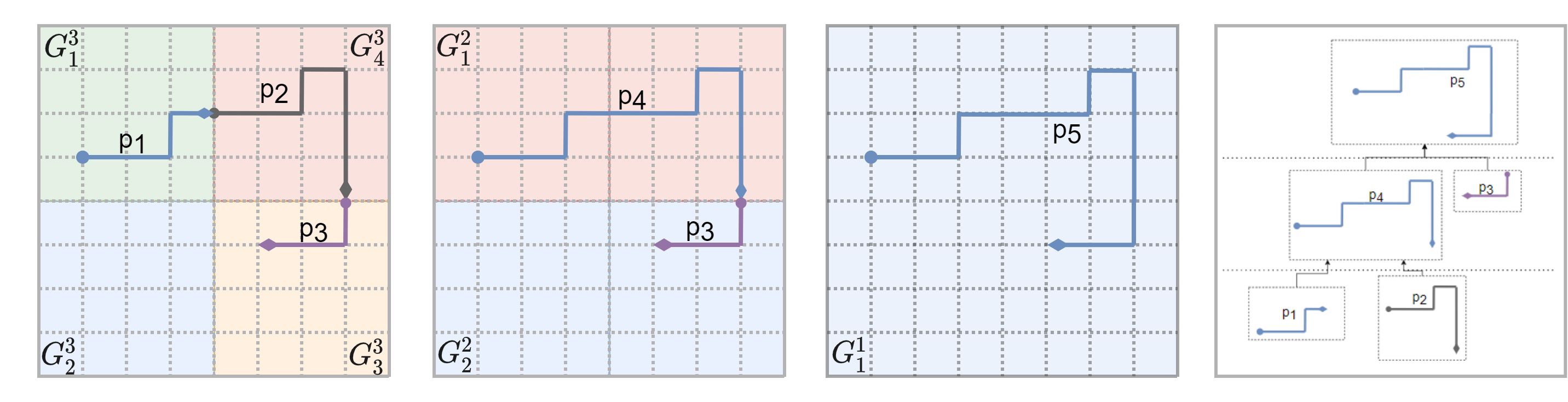}
\captionsetup{skip=1pt}
\caption{\label{fig:basic_idea_of_p2p} Illustration of the hierchical pathlet representation.  Here $G_i^k$ refers to the $i$-th region of the $k$-th layer.}
\end{figure}
\vspace{-0.5cm}

\subsection{Representing New Trajectories}
Once we obtain a set of dictionaries at multiple scales, we can use them together in representing new trajectories. We define a unified dictionary matrix  $P'$ by concatenating the dictionary matrices by column. The size of dictionary $P'$ is therefore  equal to the number of columns of $P'$. 

For any new trajectory, it can be mapped to a new representation space using  $P'$ . To be specific, representation vector is obtained by solving:
\vspace{-0.5cm}

\[ \min\limits_{{r_{i} \in \{0,1\}}} \ \sum_{i=1}^{size({P'})}|r_{i}|\ \ \ \ s.t. P'r= m\]

Here  $m$ represents the vector recording the edges covered by a new trajectory, and $r$ denotes the representation vector that we aim to solve for. This problem can be viewed as a simplified version of the original problem because the dictionary is fixed at this moment. We solve it using the same strategy described before: first compute the optimal fractional solution $r^*$ using gradient descent and then round it to get the final binary solution.
 \vspace{-0.15cm}
\section{Experiments }
 \footnotetext{Due to the space limit, details of experiments setup can be found in appendix part B.}

\subsection{Numerical Performance}
\subsubsection{The Performance Comparison with Previous Work.}
The proposed method is evaluated  on two datasets collected separately in Shenzhen \cite{zhang_urbancps_2015} and Porto  \cite{moreira-matias_predicting_2013}. 
Our research largely follows the problem formulation described in \cite{chen_pathlet_2013} but we adopt different formulation and method. In that paper, the authors first relaxed $max(R_{i,:})$ to $R_{i,j}$, and then solved it using dynamic programming, which is simple and effective. However, this relaxation operation resulted in an  redundant dictionary, providing us with room for improvement especially when $\lambda$ is small. 

\begin{table*}[]
\captionsetup{skip=3pt}
\caption{\label{tab:tab2} The performance comparison with previous work.}


\begin{tabular}{lllll|cl}
\hline
\multirow{2}{*}{Dataset}                      & \multirow{2}{*}{Method} & \multicolumn{3}{c|}{Train Phrase}                       & \multicolumn{2}{c}{Test Phrase}       \\ \cline{3-7} 
                                              &                         & Dictionary size/|T| & Representation cost & Cover ratio & Representation cost & Cover ratio     \\ \hline
\multirow{2}{*}{Porto}                        & DP                      & 1.79                & 2.14                & 100\%       & 4.19                & 88.03\%         \\
                                              & our method              & 1.02(-43.01\%)      & 2.00(-6.54\%)       & 99.6\%      & 2.75                & 93.6\%(+5.57\%) \\ \hline
\multicolumn{1}{c}{\multirow{2}{*}{Shenzhen}} & DP                      & 1.21                & 2.88                & 100\%       & 3.01                & 93.9\%          \\
\multicolumn{1}{c}{}                          & our method              & 0.91(-36.36\%)      & 2.75  (-4.51\%)     & 99.1\%      & 3.02                & 95.4\% (+1.5\%) \\ \hline
\end{tabular}
\end{table*}

In this experiment, hyperparameter $\lambda$ and $\theta$ are set as 0.1 and $\frac{1}{4}ln(2|T|)$ respectively, and we only randomly sample 3 times using strategy described before. 
As is shown in Table \ref{tab:tab2}, our approach generates a more compact and effective dictionary compared to dynamic programming methods, reducing the dictionary size by 43.01\% and 36.36\% respectively on two datasets and the representation cost is relatively lower. At the same time, it is observed that the cover ratio is very close to 1, here $\theta$ is set as $\frac{1}{4}ln(2|T|)$ instead of $ln(4|T|)$  because in the experiment we found that the method can still produce a feasible solution with low cost within 3 random sampling cycles, which further validates the effectiveness of previously derived probability bound.

\subsubsection{Reconstruction using Multi-scale Dictionary}
The hierarchical framework enables us to learn multi-level pathlet dictionaries on arbitrarily large maps and datasets with limited computational resources. In this section, we validate the above statement by comparing the performance of the dictionary directly learned on the whole map (denoted using $P_{direct}$) and the dictionaries generated by hierarchical framework on test data. Specifically, we randomly selected 10,000 trajectories from the Futian district as train set to learn the dictionary and tested it on another 10,000 trajectories. 
In Table \ref{tab:multi-scale-reconstruction }, $P_{2}$  represents  dictionaries learned on regions of the $2$-th layer and $P_{1}+P_{2}$ refers to multi-scale dictionaries.  The performance of  $P_{direct}$ can be considered as  ground truth to some extent, although it comes with significant computational resource consumption. We can observe that compared to only using  $P_{2}$, the reconstruction cost is much lower when using the multi-scale dictionary. The performance of the multi-scale dictionary is closer to that of $P_{direct}$, but consumes only 54\% of the GPU memory resources compared to the training of $P_{direct}$ and the computation time is reduced by 20\%.

\begin{table}[]
\captionsetup{skip=3pt}
\caption{\label{tab:multi-scale-reconstruction } The performance using different dictionaries.}
\begin{threeparttable}
\begin{tabular}{lcccc}
\hline
Dictionary      & Size  & \begin{tabular}[c]{@{}c@{}}Represent-\\ ation Cost\end{tabular} & \begin{tabular}[c]{@{}c@{}}GPU  \\ Memory *\end{tabular} & \begin{tabular}[c]{@{}c@{}}Running  \\ Time\end{tabular} \\ \hline
$P_{direct}$    & 13076 & 5.21                                                            & 42.8G                                                  & 1.5h                                                     \\
$P_{2}$        & 10470 & 6.05                                                            & 23.1G                                                  & 0.9h                                                     \\
$P_{1}+P_{2}$ & 12631 & 5.33                                                            & 23.1G                                                  & 1.2h                                                     \\ \hline
\end{tabular}
\begin{tablenotes}
\item[*] GPU memory here refers to the size of GPU memory needed for training instead of storage of dictionary.
\end{tablenotes}
\end{threeparttable}
\end{table}
\vspace{-0.15cm}

  \vspace{-0.1cm}
\subsection{Visualization of Pathlet Dictionary}
\begin{figure}[h]
\centering
\includegraphics[width=0.465\textwidth]{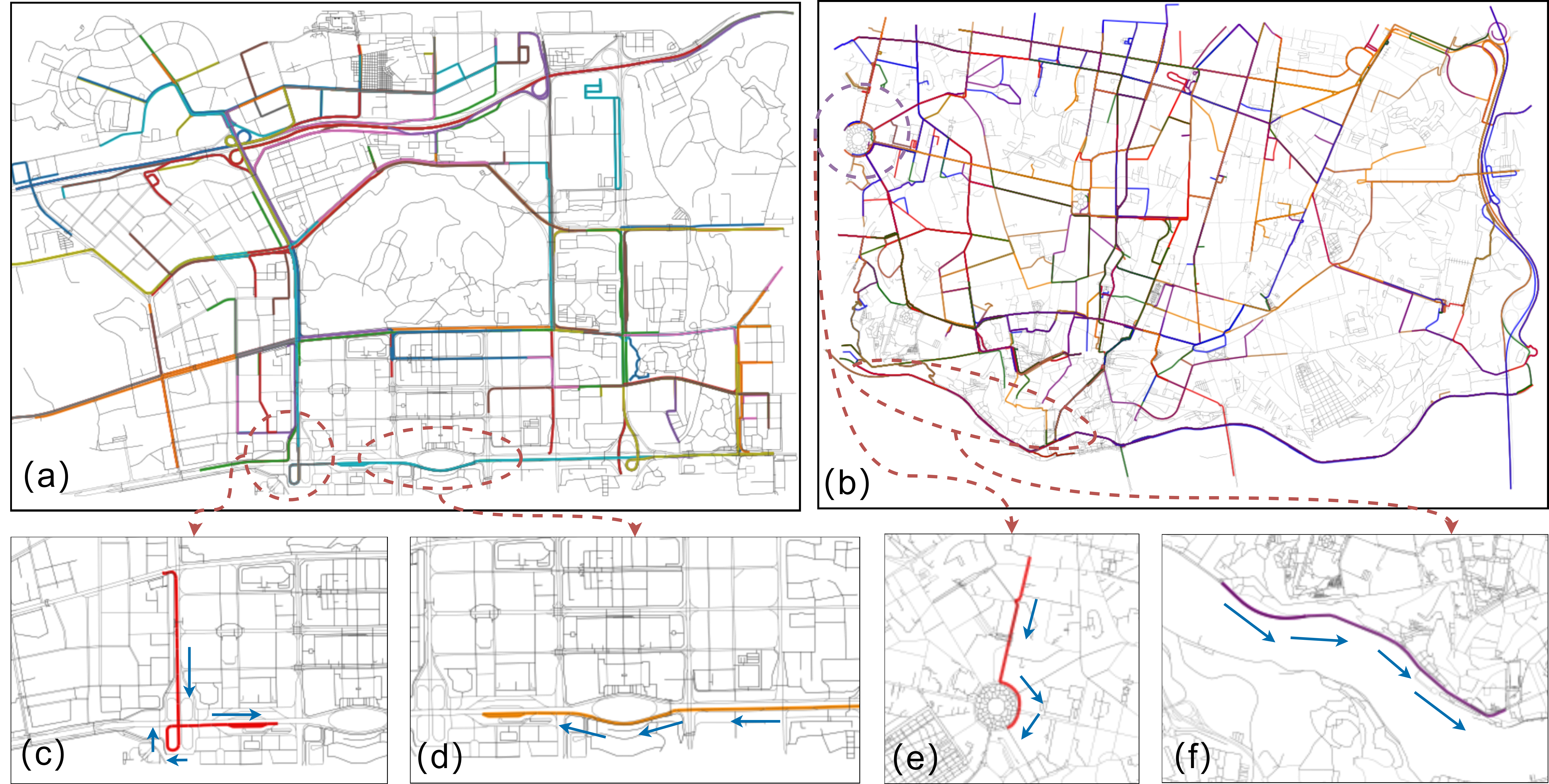}
\captionsetup{skip=6pt}
\caption{\label{fig:pathlets_extracted}Top 300 frequent pathlets in two grids.}
\end{figure}

Some frequent pathlets are visualized in Figure \ref{fig:pathlets_extracted} to intuitively verify whether the algorithm finds common mobility patterns or not.
For example, Figure \ref{fig:pathlets_extracted} (c) is a pathlet corresponding to turning left on the overpass. 
Figure \ref{fig:pathlets_extracted} (e) depicts Praça Mouzinho de Albuquerque, which is one of the famous attractions in Porto. 
These pathlets have semantic meaning consistent with our cognition in life and reveal common mobility patterns shared by numerous trajectories.
 \vspace{-0.2cm}

 \vspace{-0.05cm}
\subsection{Performance on Downstream Tasks}
\subsubsection{Application in Trip Time Prediction}


To demonstrate the effectiveness and usability of the representation vector, we utilize a simple GBDT model to predict trajectory time whose input is the combination of trajectory embedding vector and the time encoding vector. We use mean absolute error (MAE) between the predicted result and the ground truth (in seconds) as the metric to train the simple GBDT model. 
The performance of all evaluated models are summarized in the table \ref{tab:tab3}. 
It can be observed that our proposed algorithm ensures explainability of the results without compromising accuracy. One possible reason why our method outperforms others is that our vectors are naturally sparse, which makes it more robust on the test set and easier to train the model. This demonstrates the simplicity and effectiveness of our method, as well as its broad prospects in the field of application.

\vspace{-0.3cm}
\begin{table}[h]
\captionsetup{skip=5pt}
\caption{\label{tab:tab3} The performance comparison with previous work.}
\begin{tabular}{ccllll}
\hline
dataset                              & method            & MAE    & MAPE  & RMSE   & RMSLE \\ \hline
\multirow{3}{*}{Porto}               &  \cite{lam_blue_nodate} & -      & -     & -      & 0.41  \\
                                     & \cite{fu_trembr_2020}            & 171.97 & -     & -      & -     \\
                                     & Ours      & 163.9  & 26.2  & 199.74 & 0.35  \\ \hline
\multirow{2}{*}{\begin{tabular}[c]{@{}c@{}}Porto \\ (short trips)\end{tabular}} & \cite{jin_stgnn-tte_2022}         & 39.25  & 14.74 & 52.35  & -     \\
                                     & Ours      & 32.87  & 9.37  & 37.59  & 0.08  \\ \hline
\end{tabular}
\end{table}
 \vspace{-0.3cm}
\subsubsection{Application in Data Compression}
Learning a dictionary and reconstructing trajectories using elements from this dictionary can also be considered as a process of data compression. In \cite{zygouras_corridor_2018} the authors described an evaluation method based on Minimum Description Length (MDL) to measure the compression performance:
\[M D L \text { score }=\frac{\mathcal{L}(\mathcal{C})+\mathcal{L}(\mathcal{D} \mid \mathcal{C})}{\mathcal{L}(\mathcal{D})}\]
Here $\mathcal{L}(.)$ refers to the size of a data collection in bits. $\mathcal{D}$ and $\mathcal{C}$  are used to denote the dataset and the corridor set,  a concept similar to pathlets.  $\mathcal{D} \mid \mathcal{C}$ refers to   the representation of the original trajectory using corridor.
Compared to the previous method's score of 0.27 reported in \cite{zygouras_corridor_2018}, our method achieved a score of 0.21. One possible reason is our objective function and MDL score are consistent, whereas method in \cite{zygouras_corridor_2018} based on LDA does not optimize the MDL score explicitly. 
This experiment indicates that transforming  trajectories into pathlets form can effectively compress data, facilitating easier storage and transmission.


 \vspace{-0.15cm}
\section{Conclusion and Future Work}
In this study, we reformulated  the problem of pathlet learning from a collection of trajectories and solved it using a novel dictionary learning based method, resulting in a hierarchical and explainable representation of trajectories with theoretical probability bound. We tested our algorithm on two large-scale datasets. The output dictionary of pathlets provides us with deeper insight into mobility patterns. 
We also demonstrate how the pathlet could benefit downstream tasks such as trip time estimation and trajectory compression.
In future work, we will adapt our algorithm to represent trajectories in other domains, improve the numerical optimization, and further advance the theoretical analysis.

 \vspace{-0.15cm}
\section*{Acknowledgments}
This study is supported by the Tsinghua SIGS Scientific Research Start-up Fund (Grant QD2021012C).
{
\small

\bibliography{ref}

\bibliographystyle{IEEEtran}
}

\appendix
\section{Proof of Probabilistic Bound.}

\noindent\textbf{Step 1 of proof}. 
Matrix $D,R^r,M$ are binary matrix and their sizes are $|E|*|\overline{P}|$,$|\overline{P}|*|T|$,$|E|*|T|$respectively.
The constraint is $DR^r= M$, which means $(DR^r)_{i,j}$ should equal $M_{i,j}$  for all $i,j$.
$(DR^r)_{i,j}=\sum_{k=1}^{|\overline{P}|}D_{i,k}R_{k,j}^r$ is an integer greater than or equal to 0.
If $M_{i,j}=0$, then constraint will be satisfied automatically; 
For each element $M_{i,j}=1$ in $M$, the probability that constraint on $M_{i,j}$ is not satisfied is:  (which means the possibility that $e$ is not covered for  a $e\in t$)
\begin{align}
P((DR^r)_{i,j}<M_{i,j})
&=P((DR^r)_{i,j}<1)=P((DR^r)_{i,j}=0)\\
&=P(\sum_{k=1}^{|\overline{P}|}D_{i,k}R_{k,j}^r=0)
\end{align}
According to the sampling strategy, $\{R_{1,j}^r,R_{2,j}^r,...R_{|\overline{P}|,j}^r\}$ are a set of Bernoulli random variables with $P(R^r_{i,j}=1)=min(1,\theta R^*_{i,j})$. Therefore,
\begin{align}
P(\sum_{k=1}^{|\overline{P}|}D_{i,k}R_{k,j}^r=0)&=P(\sum_{k=1,D_{k,j}=1 }^{|\overline{P}|}R_{k,j}^r=0)
\\&=\prod_{k=1,D_{k,j}=1 }^{|\overline{P}|}P(R_{k,j}^r=0) \\
&=\prod_{k=1,D_{k,j}=1 }^{|\overline{P}|}(1-min(\theta R_{i,k}^*,1) )
\end{align}  
For $x\ge 0$, we have $1-min(x,1)\le e^{-x}$, therefore
\begin{align}
P(\sum_{k=1}^{|\overline{P}|}D_{i,k}R_{k,j}^r=0)
&\le \mathrm{exp} (-\theta\sum_{k=1,D_{k,j}=1}^{|\overline{P}|}R_{i,k}^*)\\
&\le \mathrm{exp} (-\theta\sum_{k=1}^{|\overline{P}|}D_{k,j}R_{i,k}^*)
\end{align}
We have $\sum_{k=1}^{|\overline{P}|}D_{k,j}R_{i,k}^*= M_{i,j}=1$ for all $i,j$ because $R^*$ satisfies the constraint $DR^*= M$.
Therefore $P((DR^r)_{i,j}<M_{i,j})\le \mathrm{exp}(-\theta )$, by the naive union bound, the possibility that the constraint is not satisfied will be less than $|T|\mathrm{exp}(-\theta)$.

\noindent\textbf{Step 2 of proof}. 
$R^r_{i,:}$ refers to the $i$ th row of $R^r$, which is a set of Bernoulli random variables $\{R_{i,1}^r,R_{i,2}^r,...R_{i,|T|}^r\}$.
${max(R^r_{i,:})}$ is also a Bernoulli random variable which means the maximum of set $R^r_{i,:}$.
We have \[P({max(R^r_{i,:})}=1)=1-P({max(R^r_{i,:})}=0)=1-\prod_{k=1}^{|T|}P({R^r_{i,k}}=0)\]
 Then the expectation of $C(R^r)$ is
\begin{align}
\mathbb{E} [C(R^r)]&= \mathbb{E} [\sum_{i=1}^{|\overline{P}|}{max(R^r_{i,:})}]
+\mathbb{E} [\lambda*\theta\sum_{i=1}^{|\overline{P}|}\sum_{j=1}^{|T|}R^r_{i,j}]\\
&=\sum_{i=1}^{|\overline{P}|}P({max(R^r_{i,:})}=1)+\lambda*\theta\sum_{i=1}^{|\overline{P}|}\sum_{j=1}^{|T|}P(R^r_{i,j}=1)\\
&\le\sum_{i=1}^{|\overline{P}|}(1-\prod_{j=1}^{|T|}(1-min(1,\theta R^*_{i,j})) )+
\lambda*\theta\sum_{i=1}^{|\overline{P}|}\sum_{j=1}^{|T|}R^*_{i,j}
\end{align}
\noindent\textbf{Lemma 1}: For a function with $|T|$ variables 
\begin{align}
f(\mathbf{x}  )=1-\prod_{i=1}^{|T|} (1-x_i)-\sum_{i=1}^{|T|} x_i 
\end{align}
$f(\mathbf{x})\le0$ when each dimension $x_i$ of $\mathbf{x}$ within [0,1].

\noindent\textbf{Proof of Lemma 1}
: First, it is easy to find that $f(\mathbf{0} )=0$, then for each dimension $x_i$, we have 
\begin{align}
\frac{\partial f}{\partial x_i} =\prod_{j!=i}(1-x_i)-1\le0
\end{align} Therefore $f(\mathbf{x})\le0$ for $\mathbf{x}$ within [0,1], QED.

Based on Lemma1, we have 
\begin{align}
\sum_{i=1}^{|\overline{P}|}(1-\prod_{j=1}^{|T|}(1-min(1,\theta R^*_{i,j})) )\le\sum_{i=1}^{|\overline{P}|}\sum_{j=1}^{|T|}min(1,\theta R^*_{i,j}).
\end{align}
Therefore, 
\begin{align}
\mathbb{E} [C(R^r)]
&\le\sum_{i=1}^{|\overline{P}|}\sum_{j=1}^{|T|}(min(1,\theta R^*_{i,j}) )+
\lambda*\theta\sum_{i=1}^{|\overline{P}|}\sum_{j=1}^{|T|}R^*_{i,j}
\\&\le
(\lambda+1)*\theta\sum_{i=1}^{|\overline{P}|}\sum_{j=1}^{|T|}R^*_{i,j}
\end{align}
On the other hand,
\begin{align}
C(R^*)&=\sum_{i=1}^{|\overline{P}|}{max( {R^*_{i,-}}) } +\lambda*\sum_{i=1}^{|\overline{P}|}\sum_{j=1}^{|T|}R^*_{i,j}\\
&>{\lambda*\sum_{i=1}^{|\overline{P}|}\sum_{j=1}^{|T|}R^*_{i,j}} 
\end{align}
Therefore, 
\begin{align}
\frac{\mathbb{E} [C(R^r)]} {C(R^*)}\le \frac{(\lambda+1)*\theta\sum_{i}\sum_{j}R^*_{i,j}}{{\lambda*\sum_{i}\sum_{j}R^*_{i,j}} }\le  \frac{(\lambda+1)*\theta}{\lambda } 
\end{align}
Therefore, $\mathbb{E} [C(R^r)]<\theta \frac{\lambda+1}{\lambda} C(R^*)$

\noindent\textbf{Step 3 of proof}. Based on the Markov inequality, we have \begin{align}
P(C(R^r)>2\mathbb{E} [C(R^r)])<1/2
\end{align} which means
\begin{align}
P[C(R^r)>2\theta \frac{\lambda+1}{\lambda}C(R^*)]&<P[C(R^r)>2\mathbb{E}[C(R^r)] ]
<1/2
\end{align}
Assume that constraint is not satisfied or  $C(R^r)>2\theta \frac{\lambda+1}{\lambda}C(R^*)$ are bad events, by the naive union bound, the probability that one of these two bad events happens is less than $1/2+ |T|\mathrm{exp}(-\theta)$. Thus, if $\theta\ge ln(2|T|)$, with positive probability there are no bad events happen and the cost of the final solution $R^r$ is  at most $2\theta \frac{\lambda+1}{\lambda}C(R^*)$. 

\section{Experiments setup}
\subsection{Dataset}
The following describes the trajectory datasets used in our study and some key statistics of datasets are summarized in Table \ref{tab:tab1}.

\noindent\textbf{Shenzhen.}  Zhang et al. \cite{zhang_urbancps_2015} released this dataset containing approximately 510k dense trajectories generated by  ~14k taxi cabs in Shenzhen, China, which can be downloaded at \cite{noauthor_httpspeoplecsrutgersedudz220datahtml_2022}.  

\noindent\textbf{Porto.} This dataset  describes trajectories performed by 442 taxis running in the city of Porto, Portugal \cite{moreira-matias_predicting_2013}. Each taxi reports its location every 15s. This dataset is used for the  Trajectory Prediction Challenge@ ECML/PKDD 2015.

\begin{figure}[h]
\centering
\includegraphics[width=0.45\textwidth]{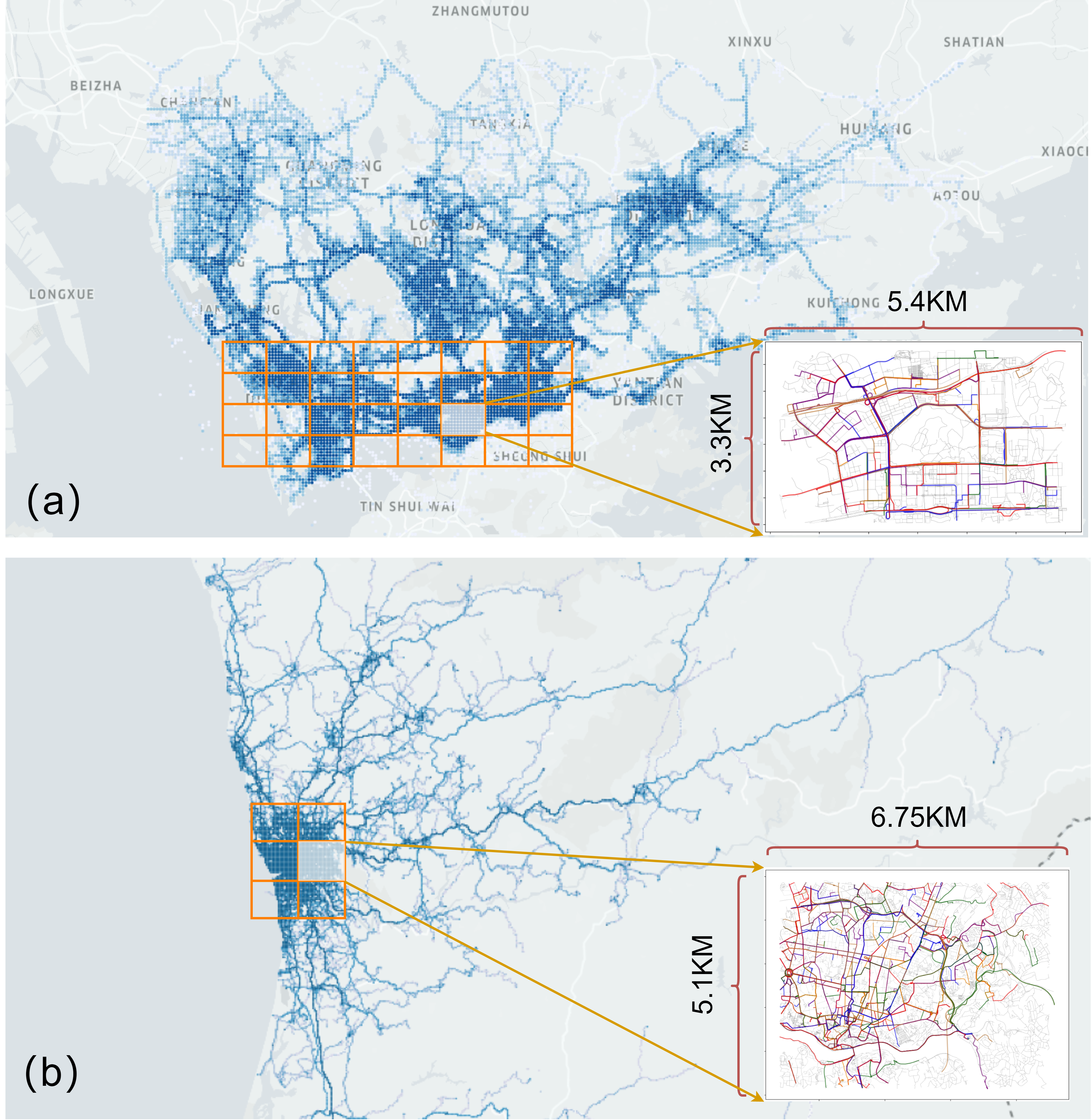}

\caption{\label{fig:dataset_info} The visualization of the real-world datasets and trajectories in L1 grid.
}
\end{figure}

Figure \ref{fig:dataset_info} displays the spatial distribution of two datasets, it can be observed that there is a significant spatial imbalance in the distribution of trajectories. In our experiments, we focus on densely populated areas. To implement our hierarchical algorithm, we first need to partition the map into smaller regions. Specifically, For the Porto dataset, we select a 15.3km x 13.5km area in the city center and divide it into six regions.
Similarly, for the Shenzhen dataset, we choose the city center area encompassing Nanshan, Futian, Bao'an, and Luohu districts, and divide it into 32 grids.

For these two datasets, We remove trajectories with less than 20 GPS sample points and use the method proposed in  \cite{meert_hmm_nodate} to convert trajectory to a series of edges on roadmap. Then the matrices $D$ and $M$ are generated as described in Section 4. 

\begin{table}[h]
\caption{\label{tab:tab1} Statistic of trajectory datasets.}
\begin{tabular}{|l|c|c|c|}
\hline
Name     & \#Trajectories & Avg.\#Points & Avg.time gap \\ \hline
Porto    & 1.2M           & 60.20        & 15.00Sec.    \\ \hline
Shenzhen & 510K           & 43.96         & 21.66Sec.    \\ \hline
\end{tabular}
\end{table}

\subsection{Evaluation Protocols and Platform}
We randomly sampled 30\% trajectories as our test dataset, and use the rest 70\% as the training dataset. 
We evaluate the quality of a resulting pathlet dictionary from the following aspects:

\begin{itemize}
    \item Size of the pathlet dictionary, i.e., the number of pathlets in the pathlet dictionary. This characterizes the compactness of a pathlet dictionary.
    \item Representation cost, i.e., the average number of pathlets used to reconstruct a trajectory, which measures the efficiency of using pathlet dictionaries to explain trajectories.

    \item Coverage Ratio, this measures whether the dictionary can cover the possible trajectories as comprehensively as possible.
\end{itemize}
Our method is implemented in Python and trained using a Nvidia A40 GPU. All experiments are run on the Ubuntu 20.04 operating system with an Intel Xeon Gold 6330 CPU.

\subsection{Pre-filtering Method}

In real-world scenarios, the number of candidate pathlets $|\overline{P}|$ is quite large, meaning that the size of the matrices $D$ and $R$ is huge. Consequently, this poses significant challenge for both computation and storage. At the same time, the pathlets we aim to identify are shared mobility patterns among multiple trajectories. Therefore, we can proactively filter out infrequent candidate pathlets without significantly impacting the results.

Specifically, for each candidate pathlet $p_i$, we traverse the trajectory dataset to count the number of trajectories that pass through $p_i$, denoted as $c_i$.
Then a threshold $c_{min}$ is set and only those candidate pathlets whose corresponding count exceeds this threshold are retained. A filtered candidate set $\overline{P}'=\{p_i\mid p_i\in \overline{P}, c_i>c_{min}\}$ can be obtained.

 In our implementation, $c_{min}$ was set as 3. To evaluate the effect of adopting pre-filtering, we randomly selected 10,000 trajectories from the Futian district and the result is shown in Table \ref{tab:filter}. We can observe that the filtering operation significantly reduces GPU memory usage and computation time, while not significantly affecting the loss.

\begin{table}[h]
\caption{The effect of adopting pre-filtering.}
\begin{tabular}{|l|c|c|c|}
\hline
Filter or not      & GPU Memory & Compute Time & Loss \\ \hline
All\_candidates      & 6.8G       & $\sim$50mins & 3112 \\ \hline
Filtered\_candidates & 2.5G       & $\sim$13mins & 3224 \\ \hline
\end{tabular}
\label{tab:filter} 
\end{table}

\subsection{Implementation Details for Trip Time Prediction Task}

\begin{figure}[h]
\centering
\includegraphics[width=0.45\textwidth]{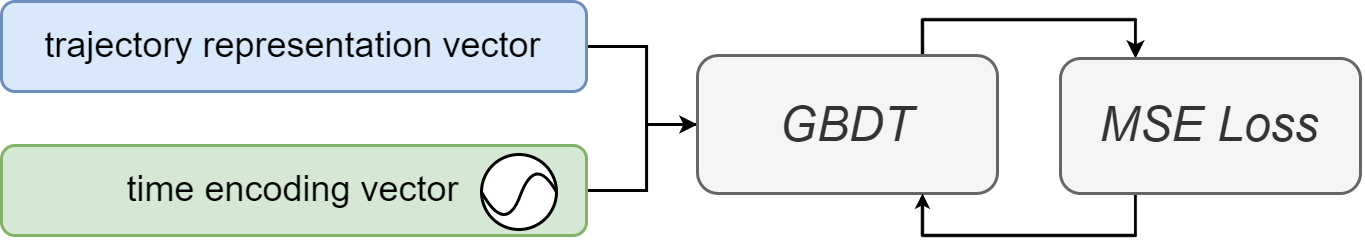}
\caption{\label{fig:GBDT4prediction}  Trip time estimation using representation vector.}
\end{figure}

 The prediction of travel time is a regression task aimed at forecasting the duration of a trajectory's journey and the result are often strongly correlated with both the chosen route and the departure time. 
 Our approach for encoding departure time in travel time prediction tasks is inspired by the positional encoding mechanism proposed in \cite{vaswani_attention_2017}. Specifically, we encode this information using sine and cosine functions and concatenate it with the trajectory representation vector as input for the GBDT model. Given the departure time $t$, the formula for the time encoding vector $TE$ is as follows:
\begin{align}
    TE(t)=sin(2\pi*\frac{t.hours*60+t.minutes}{24*60} )
\end{align}
where $t.hours$ and $t.minutes$ refer to corresponding hour and minute of $t$ respectively.
To demonstrate the effectiveness and usability of the representation vector, we utilize a simple GBDT model to predict trajectory time whose input is the combination of trajectory embedding vector and the time encoding vector. We use mean absolute error (MAE) between the predicted result and the ground truth (in seconds) as the metric to train the simple GBDT model. 
The workflow is illustrated in Figure \ref{fig:GBDT4prediction}. 
The key parameters of GBDT are set as follows: tree max depth is set as 5; number of estimators is set as 100.

To maintain fairness in the comparison, we also generated a short trip version dataset from original Porto dataset following the sampling method described in \cite{jin_stgnn-tte_2022}. We conducted algorithm testing on both of these datasets and the metrics include MAE, MAPE, RMSE and RMSLE. 

Three methods were employed for comparison:
\begin{itemize}
\item Trembr is a RNN-based framework for learning low dimensional feature vectors for trajectories, which consists of two sub-models called Traj2Vec and Road2Vec. The former encodes spatial and temporal information inherent in trajectories into embeddings, while the latter learns embeddings for road segments in road networks capturing various relationships amongst them. 
\item \cite{jin_stgnn-tte_2022} proposes a model called STGNN-TTE aiming to capture fine-grained spatial-temporal representations of traffic networks for travel time estimation. It combines spatial-temporal learning with a multi-task learning module, featuring a hybrid CNN-RNN architecture, ARIMA models, and a Transformer model.  
\item \cite{lam_blue_nodate} describes a data-driven approach for predicting taxi destination and trip time using a model ensemble of regularized linear regression with L1 penalty, gradient boosted regression trees, random forest regressor, and extremely randomized trees regressor. The ensemble predicts the final result by taking the mean of individual predictions.
\end{itemize}

\section{Supplementary Experimental Results}

\subsection{Effect of $\lambda$}
\begin{figure}[h]
\centering
\includegraphics[width=0.45\textwidth]{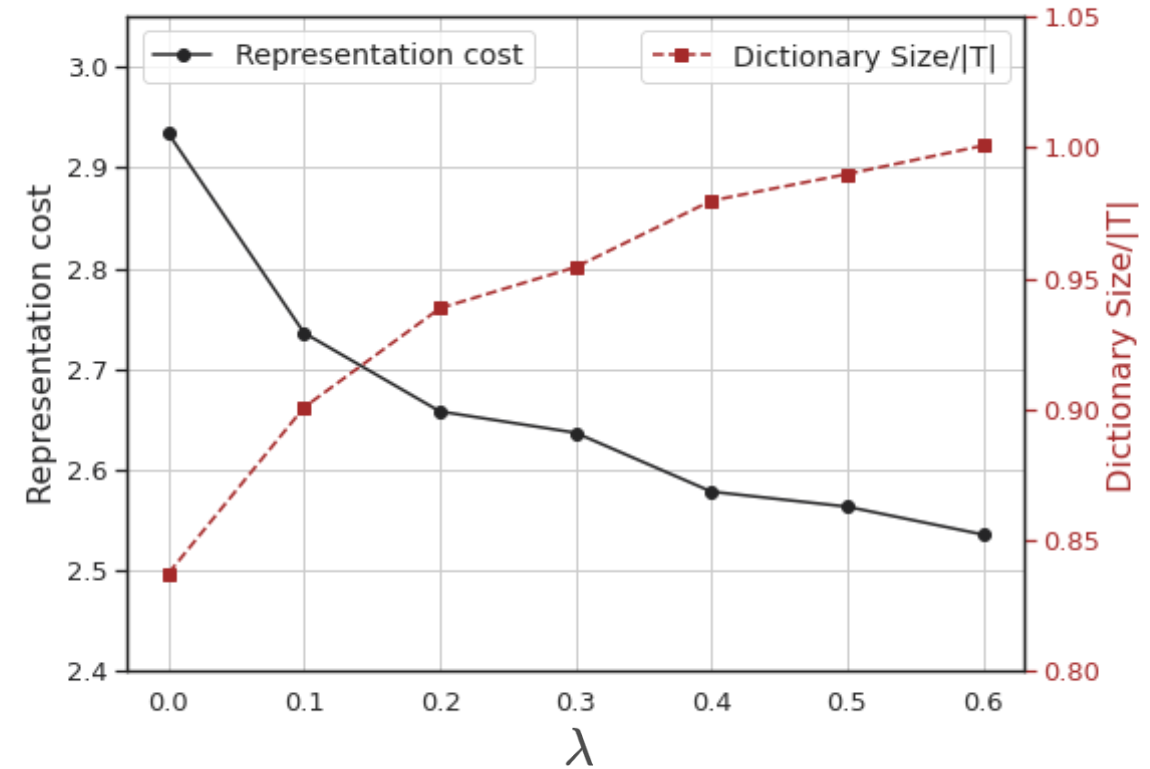}
\caption{\label{fig:effect_of_la}Effect of hyper-parameter $\lambda$.}
\end{figure}

We conducted the algorithm under different $\lambda$. It can be observed from Figure \ref{fig:effect_of_la} that when $\lambda$ increases,  the average number of pathlets need to construct a trajectory will decrease. At the same time, the size of the dictionary increases, which means that the algorithm prefers a more compact dictionary with a smaller $\lambda$. 

\subsection{Visualization of Hierarchical Pathlets}
\begin{figure*}[h]
\centering
\includegraphics[width=0.9\textwidth]{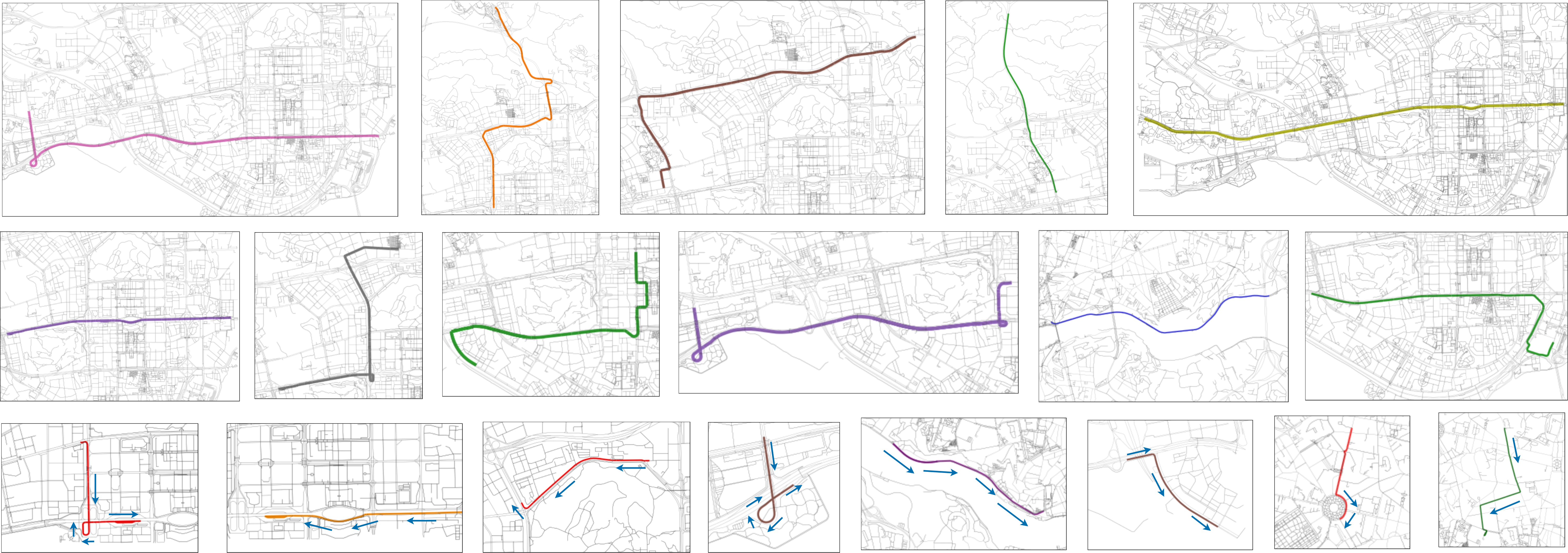}
\caption{\label{fig:heriracal_vis} The visualization of hierarchical pathlets.}
\end{figure*}

After obtaining the local pathlet dictionary, we can generate high-level pathlets based on the previous result using the method described in Section \ref{sec:Hierarchical}. 
Figure \ref{fig:heriracal_vis} shows pathlets of different levels. The three rows correspond to pathlets at different levels. Since higher-level pathlets are generated by concatenating lower-level pathlets, we can mine long-distance movement patterns from higher-level pathlets.

\subsection{Partial Reconstruction}

\begin{figure}[h]
\centering
\includegraphics[width=0.45\textwidth]{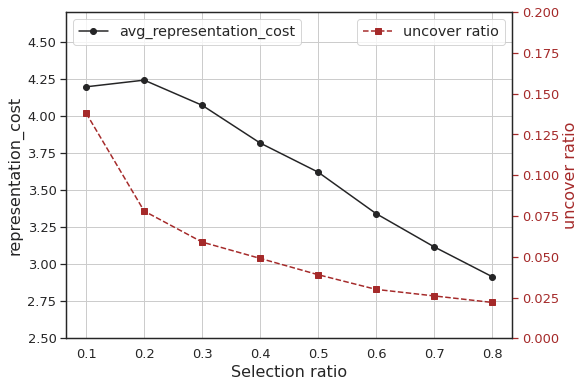}
\caption{\label{fig:part_pathlets} The representation cost and uncover ratio when varying the size of dictionary. }
\end{figure}

The  pathlet dictionary for previous evaluation is a complete dictionary that can reconstruct every trajectory in previous results. However, if we accept a small portion of trajectories that are not rebuilt, the size of the pathlet dictionary can be significantly reduced. As shown in Figure  \ref{fig:part_pathlets}, the ratio of uncover (the proportion of edges that can not be covered using pathlets from dictionary) decreases rapidly and drops below ~5\% when only 50\% of most frequent pathlets are preserved, which means the majority of trajectories still can be reconstructed using half of the pathlet dictionary.
On the other hand, it can be observed that each trajectory needs more pathlets to reconstruct itself when part of pathlets are preserved compared to representing using the complete dictionary, which reveals that there is a trade-off between redundancy and efficiency.

\section{Analysis of the Relaxation Operations in Dynamic Programming Solutions.}
\label{dp_analysis}
In \cite{chen_pathlet_2013} the authors found the challenge of problem solving in scenarios involving large-scale datasets.  To address this challenge, they proposed modifying the objective function, allowing the problem to be solved independently for each trajectory. This approach significantly reduced the complexity of the problem-solving process. Specifically, they transformed the original problem into solving a lower bound of the original problem:
\[
\begin{aligned}
 &\sum_{p \in \overline{{P}}} x_{p}+\lambda \sum_{t \in {T}} \sum_{p \in {P}(t)} x_{t, p} \\
= & \sum_{p \in \overline{{P}}} \sum_{t \in {T}(p)} \frac{x_{p}}{|{T}(p)|}+\lambda \sum_{t \in {T}} \sum_{p \in {P}(t)} x_{t, p} \\
\geq & \sum_{p \in \overline{{P}}} \sum_{t \in {T}(p)} \frac{x_{t, p}}{|{T}(p)|}+\lambda \sum_{t \in {T}} \sum_{p \in {P}(t)} x_{t, p} \\
= & \sum_{t \in {T}} \sum_{p \in {P}(t)}\left(\lambda+\frac{1}{|{T}(p)|}\right) x_{t, p}:=f
\end{aligned}
\]
 where $T (p) = \{t|t \in T , p \in P(t)\}$. The primary distinction here is the substitution of $x_p$ variable with $x_{t,p}$. However, for a specific $p \in D$, it is quite common that some $t \in {T}(p)$ do not use $p$ to represent themselves. Consequently, there exists a considerable disparity between the solution obtained by this approach and the optimal solution. Especially when lambda is small, the size of a dictionary is a crucial factor in gauging its overall quality. 

\end{document}